\documentclass[runningheads]{llncs}
\usepackage{graphicx}
\begin{document}
\title{Labeling of Multilingual Breast MRI Reports}

\titlerunning{Labeling of Multilingual Breast MRI Reports}

\author{Chen-Han Tsai\inst{1} \and
Nahum Kiryati\inst{2} \and
Eli Konen\inst{3} \and 
Miri Sklair-Levy\inst{3} \and
Arnaldo Mayer\inst{3}}

\authorrunning{C.-H Tsai et al.}

\institute{School of Electrical Engineering, Tel Aviv University, Israel \and
The Manuel and Raquel Klachky Chair of Image Processing, School of Electrical Engineering, Tel-Aviv University, Israel \and
Diagnostic Imaging, Sheba Medical Center, affiliated to the Sackler School of Medicine, Tel-Aviv University, Israel}

\maketitle              
\begin{abstract}
Medical reports are an essential medium in recording a patient's condition throughout a clinical trial. They contain valuable information that can be extracted to generate a large labeled dataset needed for the development of clinical tools. However, the majority of medical reports are stored in an unregularized format, and a trained human annotator (typically a doctor) must manually assess and label each case, resulting in an expensive and time consuming procedure. In this work, we present a framework for developing a multilingual breast MRI report classifier using a custom-built language representation called LAMBR. Our proposed method overcomes practical challenges faced in clinical settings, and we demonstrate improved performance in extracting labels from medical reports when compared with conventional approaches.

\end{abstract}

\section{Introduction}
The introduction of the Electronic Medical Record (EMR) has improved convenience in accessing and organizing medical reports. With the increasing demand for biomedical tools based on deep learning, obtaining large volumes of labeled data is essential for training an effective model. One major category where such deep learning models excel is in the area of computer assisted diagnosis (CADx), and several works (e.g.~\cite{Bien2018,gozes2020rapid,Tsai2020KneeID}) have demonstrated effective utilization of weakly labeled data to achieve promising performance. Since understanding medical data requires specialized training, datasets often contain a small subset of all past exams, that are manually relabeled by doctors for the target task. Not only is this a labour-intensive process, but the resulting dataset is often too small to represent the true distribution, resulting in underperforming models.

\begin{figure*}[]
\vskip -0.1in
\begin{center}
    \centerline{\includegraphics[width=0.9\textwidth]{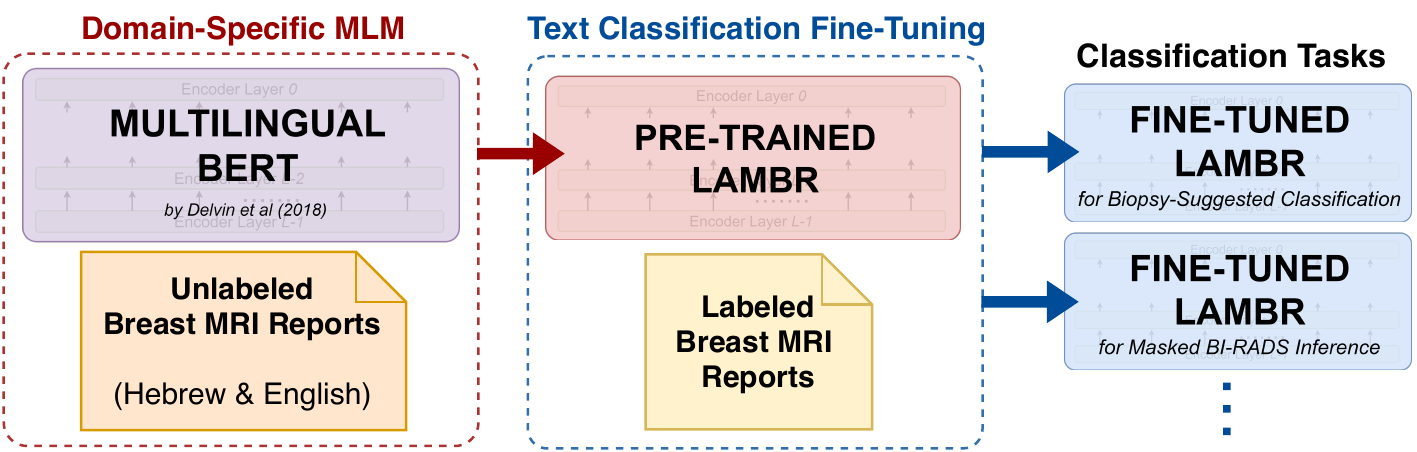}}
    \caption{An overview of training stages presented in our framework. Pre-training is performed on the multilingual BERT with unlabeled breast MRI reports to obtain a  pre-trained LAMBR. The pre-trained LAMBR is then fine-tuned using a small number of labeled reports to obtain classifiers for specific downstream text classification tasks.}
    \label{Lambert_Flowchart}
\end{center}
\vskip -0.35in
\end{figure*}

In this work, we present a framework for developing multilingual breast MRI report classifiers by using a customized language representation called LAMBR. LAMBR is first obtained by pre-training an existing language representation on a large quantity of breast MRI reports. Fine-tuning is then applied to obtain separate classifiers that can perform tasks such as: (1) determining whether the corresponding patient in the report has been suggested to undergo biopsy or (2) predicting BI-RADS \footnote{Breast Imaging-Reporting and Data System: a score between 0-6 indicating the level of severity of a breast lesion} score for the reported lesion (see Figure \ref{Lambert_Flowchart}). With such classifiers, one may avoid the manual labeling required from doctors, and instead, automatically extract a large number of weak labels from existing medical reports for training weakly supervised breast MRI CADx models. 

Prior to our work, text classification has been explored extensively by several studies such as ULMFIT~\cite{howard2018universal} and  SiATL~\cite{alex2019embarrassingly}. ELMo~\cite{DBLP:journals/corr/abs-1802-05365},  BERT~\cite{devlin2018bert}, and XLNet~\cite{yang2019xlnet} have also demonstrated adequate approaches towards text classification using the notion of a generalized language representation. However, the majority of these approaches require pre-training an encoder on a massive text corpora, and this is a time consuming and resource intensive procedure that is impractical for a clinical setting~\cite{Sharir2020TheCO}. Moreover, the majority of prior works perform encoder pre-training on widely available natural language texts which differ greatly from the scarcely available medical texts. 

\begin{figure*}[t]
\begin{center}
    \centerline{\includegraphics[width=\textwidth]{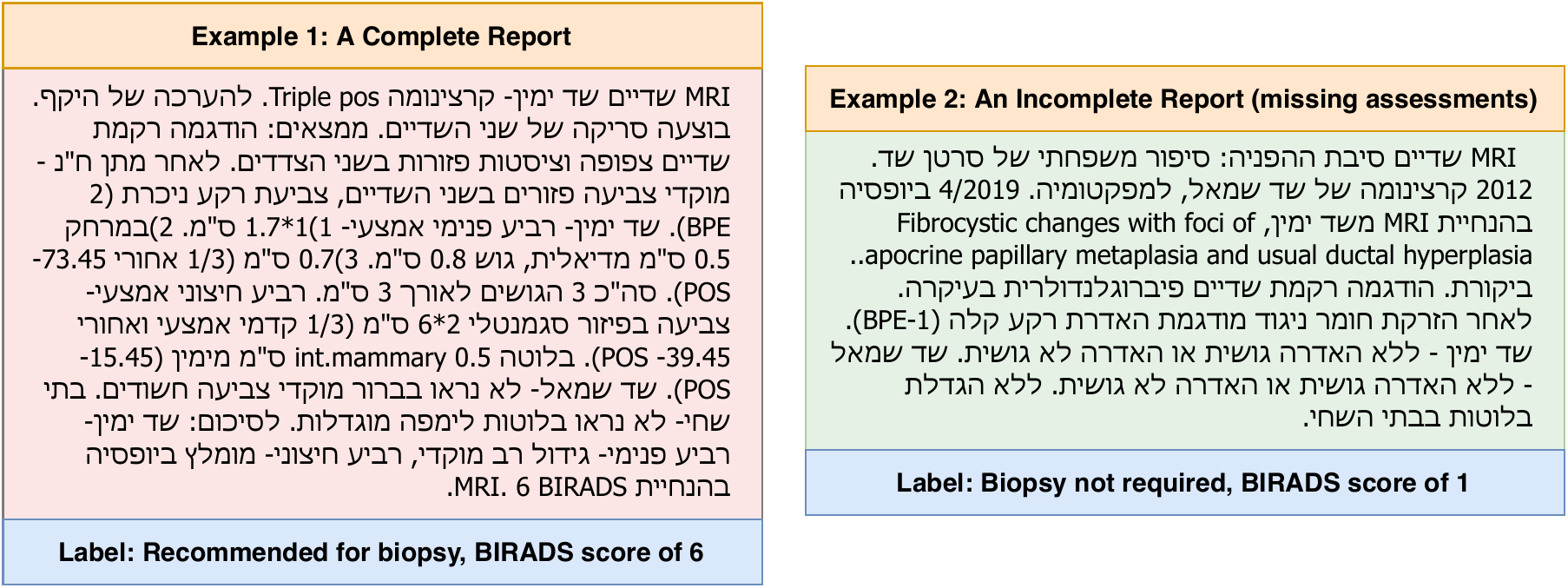}}
    \caption{Examples of breast MRI reports written in Hebrew and English (read from right to left). Example 1 is complete report parsed from the EMR, and  Example 2 is missing the final assessments due to incorrect parsing. The patient in Example 1 is recommended for biopsy, and patient in Example 2 is not required to perform biopsy.}
    \label{Lambert_Example}
\end{center}
\vskip -0.33in
\end{figure*}

To overcome the difference in distrubtion between medical texts and natural texts, BioBERT~\cite{10.1093/bioinformatics/btz682} introduced a pre-training objective that relied on a large collection of PubMed abstracts and PMC articles. Although BioBERT demonstrated improved performance compared with BERT, their method does not avoid the above resource intensive pre-training. Within English medical reports, ALARM~\cite{wood2020automated} proposed a simple approach for labeling head MRI reports by utilizing a pre-trained Bio-BERT and this avoids the expensive pre-training often required. Yet for multilingual medical reports, such Bio-BERT does not exist, and in this work, we present a solution based on the multilingual BERT. Our novel approach introduces an inexpensive pre-training objective that yields favorable text classification performance when fine-tuned, and in our experiments, we demonstrate the robustness of the resulting classifiers even in cases where parsing errors exist (see Figure \ref{Lambert_Example}). The remaining sections of our paper are organized as follows: the proposed framework is presented in Section~\ref{methods}, experimental results are reported in Section~\ref{experiments}, and the conclusion is given in Section~\ref{conclusion}.

\section{Methods}
\label{methods}

\subsection{BERT Recap}
BERT is a language representation based on the Transformer-Encoder~\cite{vaswani2017attention}. The input to the Transformer-Encoder is a sequence of tokens $\{x_i\}$ generated by WordPiece Embeddings~\cite{wu2016googles} from a given series of sentences. Special tokens are inserted and position encodings are added, and the output is a sequence of bi-directional embeddings that represents each input token~\cite{devlin2018bert}. In order to obtain the BERT language representation, Masked Language Modeling (MLM) and Next Sentence Prediction (NSP) pre-training objectives were introduced. MLM applies random masking on the 15\% of the input tokens, and BERT is trained to identify the original token of the masked token by attending to other tokens of the same sequence. The NSP objective trains BERT in understanding sentence coherence by randomly replacing the second sentence of an existing sentence pair, and BERT has to determine whether the pair are neighboring sentences.




\begin{figure*}[]
\begin{center}
    \centerline{\includegraphics[width=0.77\textwidth]{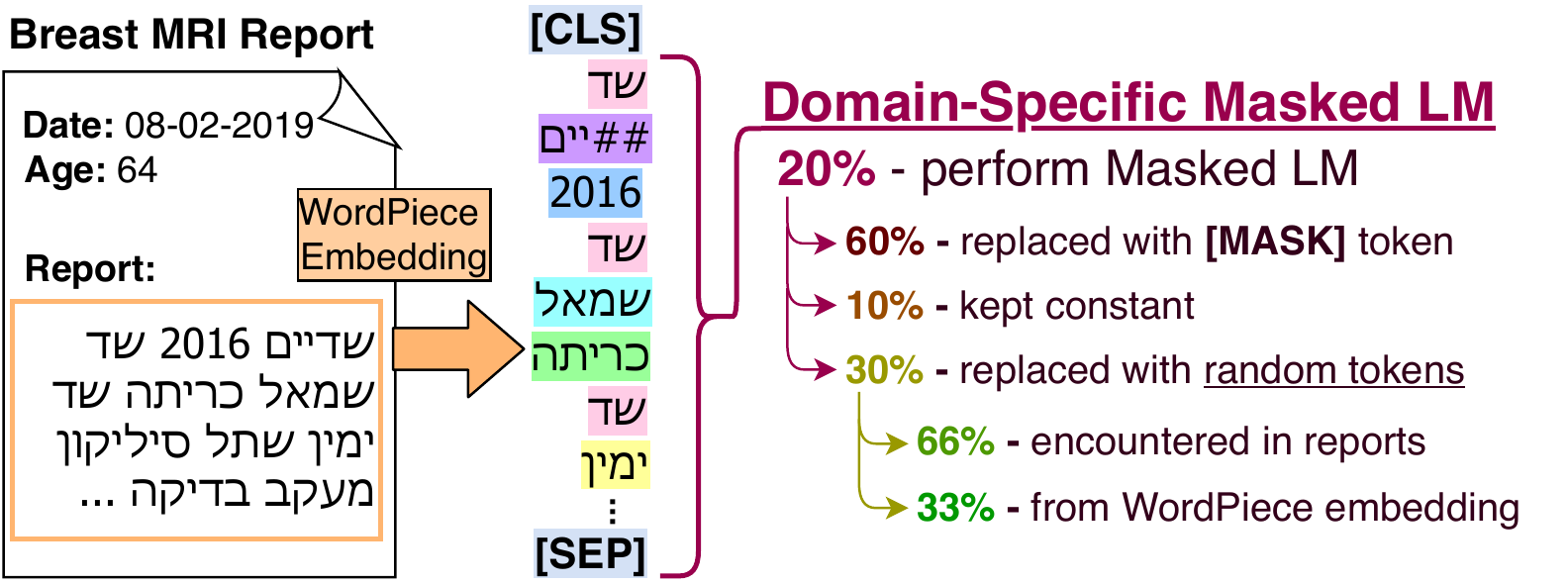}}
    \caption{An example of the tokens generated from breast MRI report using WordPiece Embeddings. During DS-MLM, a portion of the tokens are augmented and the pre-training objective is to correctly identify the original token prior to augmentation (performed only on tokens selected for augmentation).}
    \label{Lambert_DSMLM}
\end{center}
\vskip -0.4in
\end{figure*}

\subsection{Domain-Specific Masked Language Modeling}
\label{dsmlm}
The Domain-Specific Masked Language Modeling (DS-MLM) we propose is a modification of the MLM pre-training objective introduced in BERT. The multilingual BERT was trained using monolingual corpora from 104 languages, and DS-MLM aims to retrain the multilingual BERT to better model the language observed in breast MRI reports written in Hebrew and English. 
Unlike BioBert~\cite{10.1093/bioinformatics/btz682}, which relies on pre-training over massive biomedical corpora, we perform DS-MLM solely from the available breast MRI reports stored in the hospital's EMR. 

For each medical report, tokens are generated using WordPiece Embeddings (see Figure~\ref{Lambert_DSMLM}). 
The~\texttt{[CLS]} and~\texttt{[SEP]} tokens are appended to the beginning and the end of the generated tokens~(\texttt{[SEP]} tokens are not added between sentences). Since the multilingual BERT is already trained on a general domain corpora, we select 20\% of the generated tokens for MLM. Of the selected tokens, 60\% are masked using the \texttt{[MASK]} token, 30\% are replaced with existing tokens and the remaining 10\% are left unchanged. In order to expose our model to more frequent tokens observed in breast MRI reports, of the 30\% of tokens selected for replacement, two thirds are replaced with existing tokens encountered in breast MRI reports, and one third is replaced with tokens from the complete vocabulary (may include tokens corresponding to other languages). Dynamic masking is applied to allow more exposure to a broad range of tokens.

Since most medical reports contain sentences not adhering to a strict flow of ideas, we do not incorporate NSP into the pre-training objective of our framework. In addition, RoBERTa~\cite{liu2019roberta} demonstrated that the removal of NSP may even improve downstream task performance, and therefore, the  pre-training objective of the LAMBR language representation is simply DS-MLM.

\subsection{Text Classification Fine-Tuning}
\label{tcft}
\begin{figure*}[]
\begin{center}
    \centerline{\includegraphics[width=0.9\textwidth]{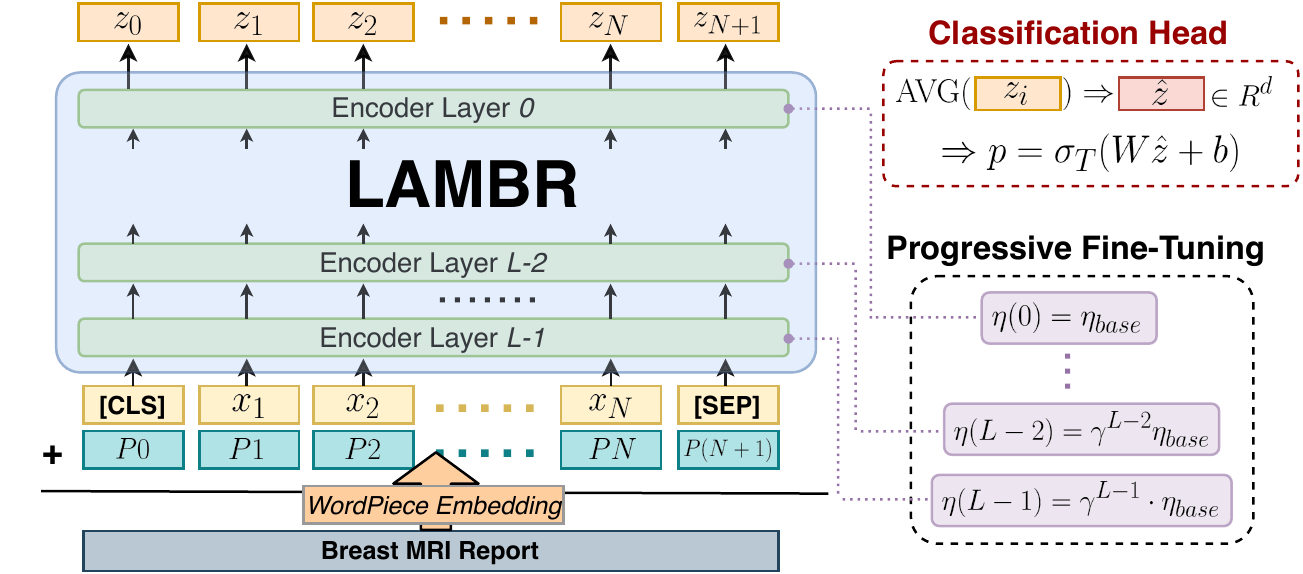}}
    \caption{An illustration of LAMBR encoding a breast MRI report. Tokens $\{x_i\}$ are generated from the report using WordPiece Embeddings, and the sum of the positional encoding and the token embedding are input to LAMBR. During TCFT, the learning rates for each layer are progressively tuned so that higher level features (layers near $0$) are updated more compared to lower-level features. The classifier head takes the average of the token embeddings $\hat z$, applies an affine transform, and passes it into a tempered Softmax ($\sigma_T$) to generate the class probabilities.}
    \label{Lambert_BERT}
\end{center}
\vskip -0.4in
\end{figure*}

Text Classification Fine-Tuning (TCFT) is a series of techniques to fine-tune a pre-trained LAMBR for performing text classification. We propose a simple classifier head to add on top of the Transformer-Encoder, and we present a method to fine-tune the complete text classifier (Transformer-Encoder along with classifier head) using a pre-trained LAMBR (see Figure~\ref{Lambert_BERT}). 

\subsubsection{Classifier Head.}
\label{classifier_head}

For a given token sequence $\{x_i\}$, we obtain the output embedding sequence $\{z_i\}$ from the pre-trained LAMBR. The average of the output token embeddings $\hat{z}$ is computed and is fed to an  affine layer which undergoes a Tempered Softmax operation ($\sigma_T$) to obtain the outputs class probabilities $p$. Namely:

\begin{equation}
    p = \sigma_T(W\hat{z} + b) \;\; \leftarrow \;\;\hat{z} = \frac{1}{N} \sum_{i=1}^N z_i  
    \label{masked_average}
\end{equation}
where $\hat{z}, b\in R^d$, $W \in R^{c \times d}$, and $p \in R^K$.


\subsubsection{Progressive Fine-Tuning.}

Inspired by \cite{yosinski2014transferable,howard2018universal}, we propose a method for fine-tuning the complete text classifier. The learning rates are adjusted such that high-level features will be updated with a higher learning rate compared to lower level features. Specifically, for a Transformer-Encoder with $L$ encoding layers $\{l_i\}_{i=0}^{L-1}$ ($l_0$ indicates the top-most layer), the layer-dependent learning rate $\eta(l)$ is formulated as: 
\begin{equation}
    \eta(l)= \eta_{base} \cdot \gamma^{l}
\end{equation}
where $\eta_{base}$ is the base learning rate and $\gamma$ is the decay factor valued between 0 and 1. Similarly, the classifier head is updated with learning rate $\eta_{base}$.

Fine-tuning is performed by optimizing the weighted Label Smoothing Loss~\cite{szegedy2015rethinking}:

\begin{equation}
    L(x,y(x)) = - \sum_{c=1}^K w(c) \cdot \Bigg[ (1-\epsilon)y_c(x) + \frac{\epsilon}{K}\Bigg] \cdot \log(p_c(x))
    \label{weighted_ce_loss}
\end{equation}
where $w(c)$ are the weights for every class $c \in K$, $\epsilon \in [0, 1)$ is the smoothing term, $y_c(x) \in \{0, 1\}$ is $1$ if $x$ belongs to class $c$, and $p_c(x)$ is the probability of $x$ belonging to class $c$ as computed in Equation \ref{masked_average}.

\section{Experiments}
\label{experiments}
In this section, we evaluate the proposed framework on two text classification tasks: (1) classifying whether the corresponding patient has been suggested to undergo biopsy and (2) predicting the BI-RADS score for the lesion reported. 

The data is a curated list of medical reports from breast MRI examinations carried out at the Sheba Medical Center, Israel. Cases that were initially diagnosed as containing potential malignant tumors have all been suggested to undergo biopsy. Breast examinations from the years 2016-2019 were involved, and a total of 10,529 medical reports were collected. Of the 10,529 breast MRI reports, 541 reports were labeled with the relevant BI-RADS score for the (single) lesion reported, and each case was labeled with whether the patient had been suggested for  biopsy.


\subsection{Training Setup}
\subsubsection{Pre-Training.}
Pre-training was performed using DS-MLM as mentioned in Section \ref{dsmlm}. Of the 9,988 reports used for pre-training, 85\% of the reports were randomly designated as the training set, and the remaining for validation. Cross Entropy loss was used for DS-MLM pre-training, and the multilingual BERT was trained for 70 epochs which took approximately 33 hours to complete using on an NVIDIA GTX 1070 8GB GPU.

\subsubsection{Biopsy-Suggested Classification}
The goal of this task to identify whether the patient in the report had been suggested to undergo biopsy or not. We perform fine-tuning as proposed in Section \ref{tcft}. Due to dataset imbalance (26.6\% of the cases were suggested for biopsy), class weights were set to the inverse of the counts per class. Evaluation was performed using 5-fold cross validation, and stratified sampling was applied to ensure equal class distribution between the training and validation sets. 

Training was performed using the Adam optimizer \cite{kingma2014adam} with a base learning rate of 1e-4 and a batch size of 8. Decay factor $\gamma$ was set to $1/4$, softmax temperature $T$ was set to 1, and the smoothing term $\epsilon$ was set to 0. The best performing model from training for 70 epochs (approx 25 mins) was evaluated.

\subsubsection{Masked BI-RADS Prediction}
Masked BI-RADS Prediction is a classification task to assign the appropriate BI-RADS score given the lesion description in the report. For reports that were parsed correctly, the BI-RADS score is written in the assessments, and a simple keyword-based tagging is often enough to label the reports with the appropriate score. However, reports might also contain BI-RADS keywords that refer to previous BI-RADS scores (for the same lesion or removed lesion), which would lead to incorrect inference if the keyword based approach was used. In addition, errors encountered during parsing would occasionally miss out sections containing the BI-RADS score, rendering the keyword-based approach useless. The text classifier we propose in our framework relies on the report descriptions alone, and is thus robust against such potential obstacles. 

Keyword search was performed on all the reports and any revealing BI-RADS scores (in the reports) were removed. This modified report was then fed into a pre-trained LAMBR for fine-tuning, and a 5-fold cross validation was performed. There were a total of 6 classes (no reports with BI-RADS score 5). Class weights were computed as the inverse of the class counts, and stratified sampling was performed to ensure equal class distribution between training and validation sets. Optimization was performed using the Adam optimizer with a base learning rate of 1e-4 and batch size of 8. The decay factor $\gamma$ was set to 1/3, Softmax temperature $T$ was set to $\sqrt{2}$, and the smoothing term $\epsilon$ was set to $1/3$. The best performing model over a training period of 70 epochs was selected for evaluation.

\begin{table*}[t]
\centering
\begin{tabular}{c|c|c|c|c}
\hline
Fine-Tuning Tasks (5-Fold Average) &  Accuracy & ROC-AUC & Macro Avg F1 & MCC    \\ \hline
Biopsy Suggested - LAMBR              & 0.965    & 0.989   & 0.935               & 0.913  \\
Biopsy Suggested - BERT               & 0.949      & 0.987     & 0.904                 & 0.8733    \\
Biopsy Suggested - Baseline           & 0.828    & -       & 0.718                   & 0.6035      \\ 
BI-RADS Prediction - LAMBR           & 0.8576   & -       & 0.7158              & 0.7594 \\
BI-RADS Prediction - BERT            & 0.795      & -       & 0.572                 & 0.672    \\ \hline 
\end{tabular}
\vspace{0.1in}
\caption{Several metrics following the five-fold cross validation for Biopsy Suggested Classification and Masked BI-RADS Prediction are presented. We compare the performance between LAMBR and BERT for both classification tasks (same classification head design, but different language representations). The baseline for Biopsy Suggested Classification is a keyword matching algorithm. Notice that in both tasks, LAMBR consistently outperforms their counterparts.}
\label{table_results}
\vskip -0.25in
\end{table*}

\subsection{Experimental Results}
The experimental results for our proposed framework are listed in Table \ref{table_results}, and we include a comparison with BERT and a baseline algorithm. 

\begin{figure*}[]
\begin{center}
    \centerline{\includegraphics[width=\textwidth]{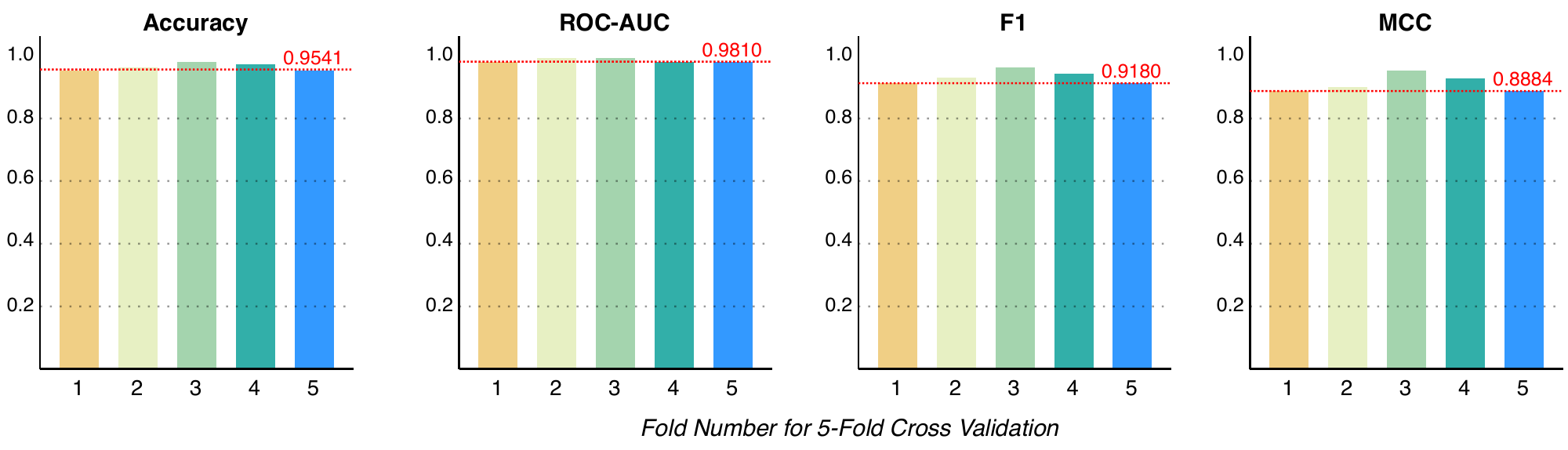}}
    \caption{Detailed visualization of the evaluation metrics for Biopsy Suggested Classification following the 5-fold cross validation. Notice the consistent performance across a 5 folds.}
    \label{Lambert_Eval}
\end{center}
\vskip -0.35in
\end{figure*}

In Biopsy Suggested Classification, the keyword matching algorithm aims to label each report in accordance with keywords that hint of a potential biopsy suggestion. Of the 541 labeled reports, 90 reports were misparsed, which contributes to a 16\% drop in accuracy. In contrast, the classifier trained and fine-tuned using our proposed framework performs consistently across all five folds (see Figure \ref{Lambert_Eval}) despite misparsed reports. We also trained a classifier with the same classification head from Section \ref{classifier_head} using a multi-lingual BERT, and we demonstrate a better classification performance with our approach.

In the task of Masked BI-RADS Prediction, the classifier trained using our framework was able to correctly predict the BI-RADS score for most of the reports. Unlike the previous task where the BI-RADS score was available, this task requires the classifier to attend to relevant context clues in the medical report for prediction (hence, the keyword-tagging algorithm does not work). An additional comparison was made between LAMBR and the pre-trained multilingual BERT, and the results in Table \ref{table_results} demonstrate a clear difference the two language representations partake in training a BI-RADS classifier.

\section{Conclusion}
\label{conclusion}

In this work, we explore the task of labeling breast MRI reports written primarily in Hebrew with occasional English texts through the use of multilingual language representations. To avoid the expensive pre-training required in obtaining a generalized language representation, the Domain-Specific Masked Language Modeling objective pre-trains a multilingual BERT on existing breast MRI reports alone to obtain the LAMBR language representation. A simple classification head is integrated onto the Transformer-Encoder, and Progressive Fine-Tuning is applied to train the classifier for its specific text classification task. 

In our experiments, we train two separate classifiers to perform two classification tasks based on breast MRI reports. In the first task, we trained a classifier to determine whether the patient described in the report has been suggested to undergo biopsy. When compared with past methods, our approach demonstrates better classification performance despite parsing errors in a portion of the reports. In the second task, we trained a classifier to predict the BI-RADS score based on the lesion description in the report. Despite the absence of the BI-RADS score in the report, our classifier was able to infer the correct BI-RADS score in the majority of the cases.



Future works may include labeling medical reports for additional pathologies written in different languages. Additional tasks (apart from text classification) such as named-entity recognition for medical reports and summary generation for biomedical texts may also be investigated. In this work, we focus on the task of medical text classification, and we believe our proposed framework may assist in generating large numbers of labels for weakly supervised training tasks required for breast MRI CADx development.

\bibliographystyle{splncs04}
\bibliography{paper6}





\end{document}